\documentclass{article}

\PassOptionsToPackage{numbers, compress}{natbib}
\usepackage[preprint]{neurips20}

\usepackage[utf8]{inputenc} % allow utf-8 input
\usepackage[T1]{fontenc}    % use 8-bit T1 fonts
\usepackage{hyperref}       % hyperlinks
\usepackage{url}            % simple URL typesetting
\usepackage{booktabs}       % professional-quality tables
\usepackage{amsfonts}       % blackboard math symbols
\usepackage{nicefrac}       % compact symbols for 1/2, etc.
\usepackage{microtype}      % microtypography
\usepackage{wrapfig}
\usepackage{graphicx}
\usepackage{subfigure}
\usepackage{lipsum}
\usepackage{footnote}
\makesavenoteenv{tabular}
\makesavenoteenv{table}

\title{Highly Available Data Parallel ML training on Mesh Networks}

\author{
Sameer Kumar, Norm Jouppi \\
Google Inc~\\
\texttt{\{sameerkm, jouppi\}@google.com}~\\
}
\date{October 2020}

\begin{document}

\maketitle

\begin{abstract}
Data parallel ML models can take several days or weeks to train on several accelerators. The long duration of training relies on the cluster of resources to be available for the job to keep running for the entire duration. On a mesh network this is challenging because failures will create holes in the mesh. Packets must be routed around the failed chips for full connectivity. In this paper, we present techniques to route gradient summation allreduce traffic around failed chips on 2-D meshes. We evaluate performance of our fault tolerant allreduce techniques via the MLPerf-v0.7 ResNet-50 and BERT benchmarks. Performance results show minimal impact to training throughput on 512 and 1024 TPU-v3 chips.  
\end{abstract}

\section{Introduction}

Deep learning has found a wide range of applications from image classification~\cite{DBLP:journals/corr/HeZRS15}, object detection~\cite{DBLP:journals/corr/HeGDG17},  language modeling~\cite{DBLP:journals/corr/VaswaniSPUJGKP17, devlin2018bert}, content recommendation~\cite{NIPS2013_5004}, speech recognition~\cite{amodei2015deep}, reinforcement models for gaming and self driving cars. To enable high quality, the models are trained on large datasets typically for several days on tens to hundreds of accelerators such as NVIDIA GPUs.  The popular algorithm for distributed training is mini-batch data parallel training~\cite{li2020pytorch}. Here each worker executes ML training forward and backward passes on a mini-batch. The computed gradients from the loss function are then summed globally via an allreduce operation. 

Large scale ML data parallel training relies on a scalable global allreduce library optimized for the training platform. As several training steps are typically executed with the same batch size over several days, the same number of accelerators must be dedicated to the training job. Typically, in a datacenter cluster, when a failure happens the job will restart from a recent checkpoint and the failed server is swapped with a spare server on the data center network. This approach works well on a fully connected data center network. 

On a 2-D mesh network, when there is a failure, any of the following approaches could be used:

\begin{itemize}
    \item Fire Fighter approach: here data center specialists or even robots can quickly go and repair the failed host and make all servers in the mesh available for the job.
    \item Sub-mesh jobs: the ML training job is executed on a mesh smaller than the original mesh. If the failure is in the middle of the mesh, the training job may only execute on half the mesh resulting in significant loss to training throughput in that job, while the unavailable servers are repaired.
    \item Rebuild mesh with hot spares~\cite{cerebras_wafer_Scale}:  in this case, when there is a failure, the mesh data network is rebuilt via the use of spares. Note, there is additional cost to having spares in rows and columns of the mesh. 
    \item Fault tolerant technique: here there are no additional spares and network packets are routed around the failed nodes in the mesh network.  The main challenge here is to execute the gradient global summation efficiently on the entire mesh even with failed chips.
\end{itemize}

We present techniques to execute fast global summation operations on 2-D meshes with failed chips. We present performance results on the Google TPU-v3 machine~\cite{Jouppi_17, cloud_tpu, kumar2019scale}. The next section describes optimized allreduce algorithms on 2-D meshes with failures.  

\section{Mesh Algorithms}

\begin{figure}
    \centering
    \begin{minipage}{0.475\textwidth}
        \centering
        \includegraphics[width=0.9\textwidth]{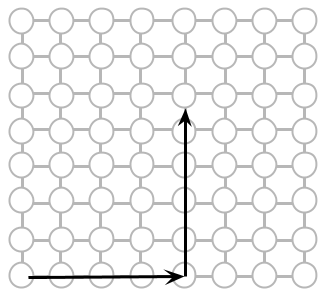}
        \caption{Dimension order routing on 2-D meshes.}
        \label{fig:dim_order}
    \end{minipage}\hfill
    \begin{minipage}{0.475\textwidth}
        \centering
        \includegraphics[width=0.8\textwidth]{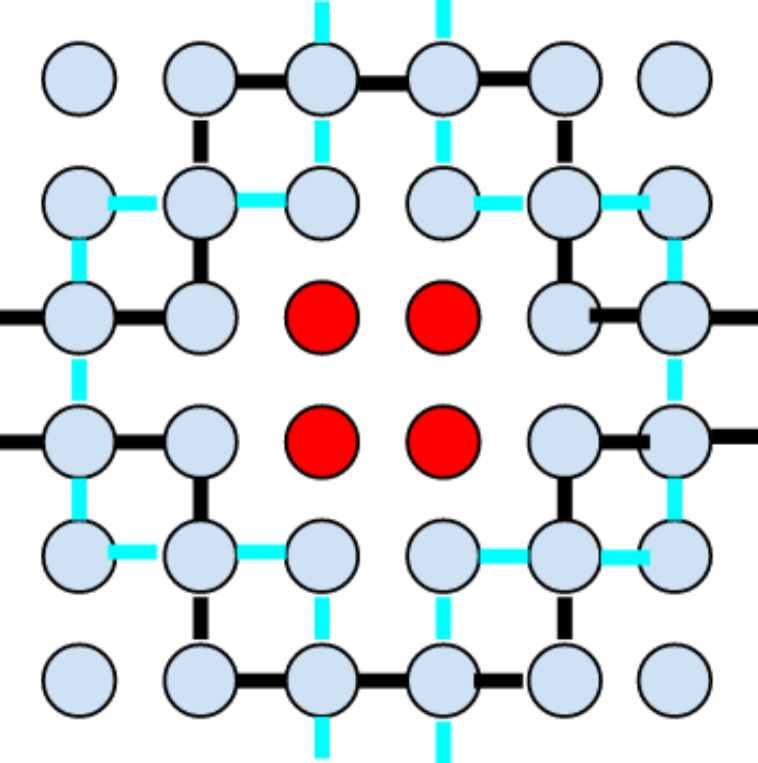}
        \caption{Non-minimal routing on a 2-D mesh with a 2x2 failed region of chips.}
        \label{fig:route_around}
    \end{minipage}
\end{figure}

Routing on the TPU-v3 mesh communication network uses standard dimension order routing (Figure~\ref{fig:dim_order}). To increase availability of TPU meshes, we explore a mode where we enable high throughput allreduce with failed regions that are 2x2 or 4x2 blocks of chips. These correspond to a single board or 2 boards on a single host of the TPU-v3 machine.  In the presence of a failed contiguous region of chips the network routing would use non-minimal paths around the failed region as shown in Figure~\ref{fig:route_around}. As long as the non-minimal paths don’t form cycles, significant additional Virtual channel resources are not required on a 2D mesh~\cite{on_chip_fault_tolerant_routing, on_chip_adaptive_routing}.

\subsection{Allreduce on 2-D mesh networks}

\begin{figure}
    \centering
    \begin{minipage}{0.475\textwidth}
        \centering
        \includegraphics[width=.9\textwidth]{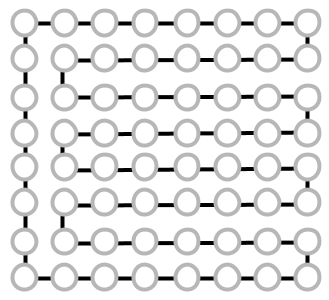}
        \caption{1-D algorithm for building a near-neighbor Hamiltonian ring on a 2-D mesh.}
        \label{fig:1-d-ring}
    \end{minipage}\hfill
    \begin{minipage}{0.475\textwidth}
        \centering
        \includegraphics[width=.9\textwidth]{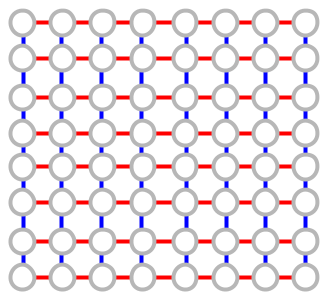}
        \caption{2-D algorithm for allreduce on 2-D meshes. Here, there are two concurrent reductions colored red and blue}
        \label{fig:2-d-ring}
    \end{minipage}
\end{figure}

\begin{wrapfigure}{r}{0.5\textwidth}
    \centering
    %% \captionsetup{format=plain}
    \includegraphics[width=.85in]{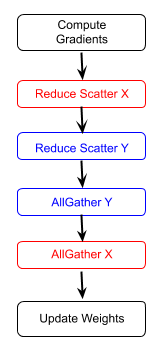}
    \caption{Steps in the data parallel gradient summation allreduce operation on a 2-D mesh.}
    \label{fig:flowchart}
\end{wrapfigure}

The ring algorithm~\cite{xin_ring} is widely used when implementing the allreduce collective operation. For example, the NVIDIA NCCL~\cite{nccl} uses a ring algorithm for reductions between GPUs over the NVLink network~\cite{nvlink_nvswitch}. Ring reductions schemes for allreduce on a 2-D mesh can use either the 1-D algorithm (Figure~\ref{fig:1-d-ring}) or the 2-D algorithm~\cite{jain_11_bucket} as shown in Figure~\ref{fig:2-d-ring}. In the 1-D scheme, a Hamiltonian circuit is built such that nodes only communicate with a downstream and upstream near neighbor on the 2-D mesh. This scheme can have a high latency of $O(N^2)$ store-forward transfers on an $N\times N$ mesh. This may be significant for short and medium sized transfers. In the 2-D algorithm (Figures~\ref{fig:2-d-ring} and~\ref{fig:flowchart}), nodes execute allreduce rings along the X dimension first (shown in red) and then along the y dimension (shown in blue). After the first phase along the X dimension of an $N\times N$ mesh, each node has a reduced shard of size $1/N$ the total allreduce payload. In the second phase, the small summed shard is summed along the Y dimension to produce a shard of size $1/N^2$ the allreduce payload. The result shard is broadcast over two gather phases on the Y and then the X dimensions.  Note, the 2-D algorithm has a lower latency of O(N) on an $N\times N$ mesh.  For full throughput, we can execute two concurrent flips over half the payload along X and Y dimensions and then execute the second phase along Y and X dimensions, respectively. This results in twice the throughput in the 2-D algorithm. 

\begin{figure}[h]
    \centering
    \begin{minipage}{0.475\textwidth}
        \centering
        \includegraphics[width=.9\textwidth]{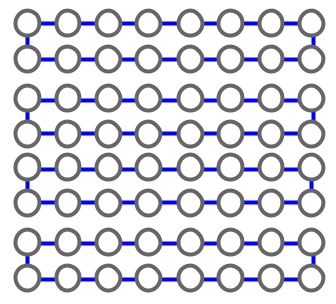}
        \caption{First phase in the alternate 2-D allreduce scheme, where pairs of two rows form a ring and execute a ring allreduce.}
        \label{fig:alt_2d_p0}
    \end{minipage}\hfill
    \begin{minipage}{0.475\textwidth}
        \centering
        \includegraphics[width=.9\textwidth]{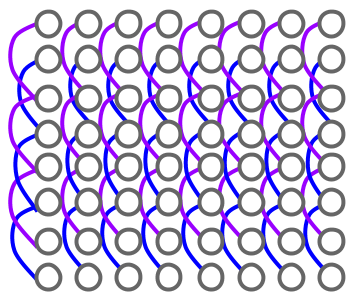}
        \caption{Second phase of alternate 2-D allreduce scheme, where nodes in alternate rows form a ring.}
        \label{fig:alt_2d_p1}
    \end{minipage}
\end{figure}

A possible downside of the 2-D scheme is that links are shared by traffic in two directions on a 2-D mesh resulting in network contention. An alternative scheme that does not require multiple colors (concurrent flips) is presented in Figures~\ref{fig:alt_2d_p0} and~\ref{fig:alt_2d_p1}. Note, in this scheme we build rings of size $2\times N$ nodes.  As each link is only used by one all-reduce ring, this scheme can achieve high link throughput in the first phase.  However, note in the second phase (Figure~\ref{fig:alt_2d_p1}) nodes must communicate with ring neighbors that skip rows and that may result in some network congestion. However, on large meshes the communication volume is significantly reduced in the second phase and this phase will not significantly impact the throughput of the allreduce operation. 

\subsection{Fault tolerant allreduce schemes}

\begin{figure}
    \centering
    \begin{minipage}{0.475\textwidth}
        \centering
        \includegraphics[width=.8\textwidth]{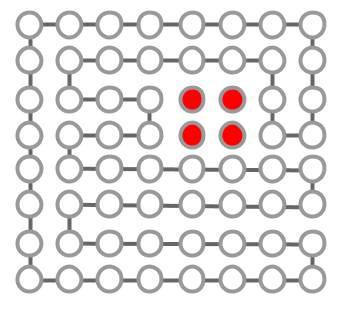}
        \caption{1-D scheme to build an allreduce ring on a 2-D mesh with a 2x2 failed region. Failed chips marked in red}
        \label{fig:ha-1d}
    \end{minipage}\hfill
    \begin{minipage}{0.475\textwidth}
        \centering
        \includegraphics[width=.85\textwidth]{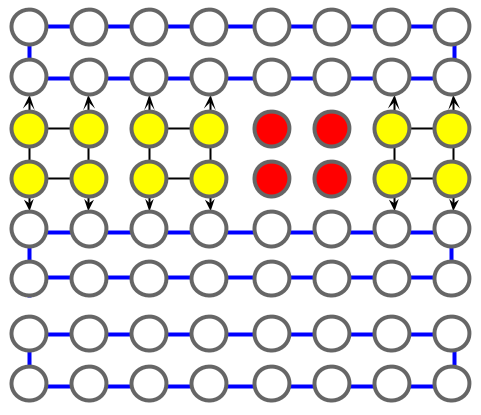}
        \caption{Fault tolerant allreduce rings built using a 2-D algorithm. The failed chips are marked in red and the peers of failed chips marked in yellow forward data to the full blue rings.}
        \label{fig:ha-2d}
    \end{minipage}
\end{figure}

We next present algorithms for allreduce when there are failures. Figure~\ref{fig:ha-1d} shows the 1-D scheme on a 2-D mesh with a contiguous failed region of size 2x2. Note, the 1-D Hamiltonian circuit can be built when the failed chips are form a contiguous region that is of even size and starts on even rows and columns. 

\begin{wrapfigure}{r}{0.5\textwidth}
    \centering
    %% \captionsetup{format=plain}
    \includegraphics[width=2.6in]{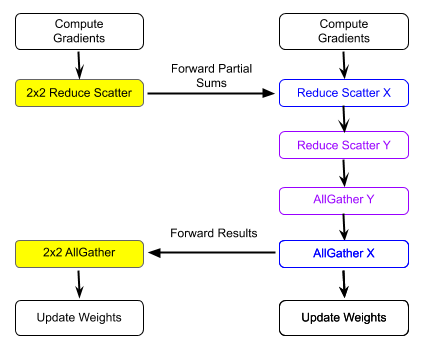}
    \caption{Steps in the forwarding scheme with a failed 2x2 region on a 2-D mesh.}
    \label{fig:ha-flowchart}
\end{wrapfigure}

When the shape of the failed region is 2kx2 or 2x2k we can build optimal 2-D allreduce rings as shown in Figure~\ref{fig:ha-2d}. Here we build rings on nodes along two consecutive rows along the X dimension similar to the allreduce scheme in Figures~\ref{fig:alt_2d_p0} and~\ref{fig:alt_2d_p1}. Neighbors of the failed chips (shown in yellow) form smaller rings in similar 2x2 blocks as shown in Figure~\ref{fig:ha-2d}. After a ring reduction round, the partial sums on yellow nodes are forwarded to neighbors on full blue rings as shown in Figures~\ref{fig:ha-2d} and~\ref{fig:ha-flowchart}. In the first phase of the allreduce, the blue rings do not share network links and that results in high throughput. A similar forwarding scheme can be used in the second phase, where nodes that are Y neighbors of the failed chips forward their contributions to Y  neighbors on columns that don’t have failed chips. However, for simplicity, we just use the route around scheme shown in Figure~\ref{fig:route_around} to execute ring reductions in the second phase. We find the route around scheme works quite well as the second phase transfers 1/2N less payload than the first phase.

\section{Experiments}
We compare the performance of fault tolerant 2-D allreduce with standard 2-D allreduce on the Google TPU-v3 machine. We use the MLPerf-v0.7~\cite{mlperf} ResNet-50 and BERT benchmarks to compare both schemes. These benchmarks are developed in the TensorFlow~\cite{abadi2016tensorflow} programming framework. ResNet-50~\cite{DBLP:journals/corr/HeZRS15, resnet_18} is an image classification model that is one of the most widely-used models for ML benchmarking. The MLPerf~\cite{mlperf} ResNet-50 benchmark trains the model on the ImageNet-1K~\cite{ILSVRC15} dataset. The BERT~\cite{devlin2018bert} model is a pre-training task for language understanding with a bi-directional transformer architecture that trains on the Wikipedia dataset. For a fine grained analysis of allreduce we disable the weight update sharding~\cite{xu2020automatic} technique in the XLA compilers for TPUs~\cite{xla}. Note the weight update sharding technique distributes the optimizer weight updates by executing them on the partially summed shards produced during the ring-allreduce algorithm. 

Table~\ref{tab:e2e_times} shows the end to end time with the two MLPerf benchmarks on 512 and 1024 TPU chips that had 16x32 and 32x32 mesh topologies. The failed region here has a shape of 4x2 with 8 total failed chips. Note, the run-to-run variance here is under 2\%. The table also shows the relative efficiency of fault tolerant vs full meshes. The relative efficiency also compensates for the reduction in the number of chips in addition to overheads from the fault tolerant allreduce scheme. From the table we can conclude the maximum overhead from fault tolerant allreduce is 5.4\%. On 512 chips, the fault tolerant job is more efficient than the full mesh job. This may be because the fault tolerant training job results in better regularization than the full mesh job. Table~\ref{tab:comm_overhead} compares the communication overheads from standard allreduce vs fault tolerant allreduce.

\begin{table}[h]
  \centering
  \begin{tabular}{l|cc|cc|c}
    \toprule            
Benchmark & \multicolumn{2}{c|}{Full Mesh} & \multicolumn{2}{c|}{Fault Tolerant Mesh} & Relative Efficiency \\
& TPU Chips & Benchmark Time & TPU Chips & Benchmark Time & \\
\midrule
ResNet-50 & 512 & 1.80 minutes &  504 & 1.84 minutes & 0.99 \\
ResNet-50 & 1024 & 1.08 minutes & 1016 & 1.15 minutes & 0.946 \\
BERT & 512 & 1.90 minutes & 504 & 1.92 minutes & 1.02 \\
BERT & 1024 & 1.16 minutes & 1016 & 1.19 minutes & 0.986 \\
    \bottomrule
  \end{tabular}
   \caption{End to end times from the MLPerf-v0.7 benchmarks are shown on both the standard 2-D mesh and the fault tolerant 2-D mesh with a failed 4x2 region. The relative efficiency shows the performance degradation from the fault tolerant mesh.}
  \label{tab:e2e_times}
\end{table}  

\begin{table}[h]
  \centering
  \begin{tabular}{l|cc|cc}
    \toprule            
Benchmark & \multicolumn{2}{c|}{Full Mesh} & \multicolumn{2}{c|}{Fault Tolerant Mesh}  \\
& TPU Chips & Allreduce overhead & TPU Chips & Allreduce Overhead \\
\midrule
ResNet-50 & 512 & 4.2\% &  504 &  6.4\% \\
ResNet-50 & 1024 & 8.8\%  & 1016 & 11\%  \\
BERT & 512 & 3.7\% & 504 & 4.7\%  \\
BERT & 1024 & 6.0\% & 1016 & 7.8\% \\
    \bottomrule
  \end{tabular}
   \caption{Here we show the communication overhead percent in the device execution step time.}
  \label{tab:comm_overhead}
\end{table}  

\section{Summary}
We presented a fault tolerant allreduce algorithm on 2-D mesh networks. The fault tolerant allreduce algorithm enabled higher availability as ML training jobs could execute with a failed region of up to 8 chips (in a contiguous 4x2 topology). This resulted in ML training running on 504 of the 512 chips and 1016 out of 1024 chips, respectively. In the MLPerf BERT and ResNet-50 benchmarks, the training time was minimally affected with under 6\% overheads in the worst case. This scheme has been available in production on Google data centers for training of Google ML models. Techniques presented are general and the fault tolerant schemes could be extended to other architectures that have 2-D meshes. 

In future, we plan to  implement the weight update sharding optimization~\cite{xu2020automatic} on meshes with failures. As the fault tolerant allreduce algorithm builds reduce-scatter and all-gather rings on complete dimensions, the optimizer weight updates can be computed at the end of the reduce-scatter phase and the updated weights can be forwarded to the nodes that are neighbors of the failed chips and do not participate in those allreduce rings.

The schemes presented in this paper can be extended to n-D mesh and n-D torus networks. On such networks multi-dimensional rings will need to be built to reach peak performance. The main challenge would be to ensure efficient forwarding of partial sums from neighbors of failed chips to the full allreduce rings.

\section{Acknowledgements}

We would like to thank Bjarke Roune, Jeremy Lau, Steve Lacy, Andy Swing, Peter Gavin, Blake Hechtman, David Manemer, Robert Hundt, Naveen Kumar, Cliff Young, George Kurian, Peter Brandt and Pankaj Kanwar for their support with TPU compilers, systems and performance analysis.  

\bibliography{bibliography}{}

\begin{thebibliography}{23}
\providecommand{\natexlab}[1]{#1}
\providecommand{\url}[1]{\texttt{#1}}
\expandafter\ifx\csname urlstyle\endcsname\relax
  \providecommand{\doi}[1]{doi: #1}\else
  \providecommand{\doi}{doi: \begingroup \urlstyle{rm}\Url}\fi

\bibitem[clo()]{cloud_tpu}
{Empowering businesses with Google Cloud AI}.
\newblock \url{https://cloud.google.com/tpu}.

\bibitem[mlp()]{mlperf}
Mlperf: Fair and useful benchmarks for measuring training and inference
  performance of ml hardware, software, and services.
\newblock \url{http://mlperf.org}.

\bibitem[ncc()]{nccl}
{NVIDIA Collective Communications Library (NCCL)}.
\newblock \url{https://developer.nvidia.com/nccl}.

\bibitem[nvl()]{nvlink_nvswitch}
{}nvlink and nvswitch: The building blocks of advanced multi-gpu communication.

\bibitem[xin()]{xin_ring}
{}bandwidth optimal all-reduce algorithms for clusters of workstations.

\bibitem[xla()]{xla}
Xla: Optimizing compiler for tensorflow.
\newblock \url{https://www.tensorflow.org/xla}.

\bibitem[cer(2019)]{cerebras_wafer_Scale}
Wafer scale deep learning.
\newblock \emph{Cerebras tutorial at Hotchips'19}, 2019.
\newblock URL
  \url{https://www.hotchips.org/hc31/HC31_1.13_Cerebras.SeanLie.v02.pdf}.

\bibitem[Abadi et~al.(2016)Abadi, Barham, Chen, Chen, Davis, Dean, Devin,
  Ghemawat, Irving, Isard, et~al.]{abadi2016tensorflow}
Mart{\'\i}n Abadi, Paul Barham, Jianmin Chen, Zhifeng Chen, Andy Davis, Jeffrey
  Dean, Matthieu Devin, Sanjay Ghemawat, Geoffrey Irving, Michael Isard, et~al.
\newblock {TensorFlow}: {A} system for large-scale machine learning.
\newblock In \emph{12th $\{$USENIX$\}$ symposium on operating systems design
  and implementation ($\{$OSDI$\}$ 16)}, pages 265--283, 2016.

\bibitem[Amodei et~al.(2015)Amodei, Anubhai, Battenberg, Case, Casper,
  Catanzaro, Chen, Chrzanowski, Coates, Diamos, Elsen, Engel, Fan, Fougner,
  Han, Hannun, Jun, LeGresley, Lin, Narang, Ng, Ozair, Prenger, Raiman,
  Satheesh, Seetapun, Sengupta, Wang, Wang, Wang, Xiao, Yogatama, Zhan, and
  Zhu]{amodei2015deep}
Dario Amodei, Rishita Anubhai, Eric Battenberg, Carl Case, Jared Casper, Bryan
  Catanzaro, Jingdong Chen, Mike Chrzanowski, Adam Coates, Greg Diamos, Erich
  Elsen, Jesse Engel, Linxi Fan, Christopher Fougner, Tony Han, Awni Hannun,
  Billy Jun, Patrick LeGresley, Libby Lin, Sharan Narang, Andrew Ng, Sherjil
  Ozair, Ryan Prenger, Jonathan Raiman, Sanjeev Satheesh, David Seetapun,
  Shubho Sengupta, Yi~Wang, Zhiqian Wang, Chong Wang, Bo~Xiao, Dani Yogatama,
  Jun Zhan, and Zhenyao Zhu.
\newblock Deep speech 2: End-to-end speech recognition in english and mandarin,
  2015.

\bibitem[Devlin et~al.(2018)Devlin, Chang, Lee, and Toutanova]{devlin2018bert}
Jacob Devlin, Ming-Wei Chang, Kenton Lee, and Kristina Toutanova.
\newblock {Bert}: Pre-training of deep bidirectional transformers for language
  understanding.
\newblock \emph{arXiv preprint arXiv:1810.04805}, 2018.

\bibitem[{Ebrahimi} et~al.(2012){Ebrahimi}, {Daneshtalab}, {Liljeberg},
  {Plosila}, and {Tenhunen}]{on_chip_adaptive_routing}
M.~{Ebrahimi}, M.~{Daneshtalab}, P.~{Liljeberg}, J.~{Plosila}, and
  H.~{Tenhunen}.
\newblock Lear -- a low-weight and highly adaptive routing method for
  distributing congestions in on-chip networks.
\newblock In \emph{2012 20th Euromicro International Conference on Parallel,
  Distributed and Network-based Processing}, pages 520--524, 2012.

\bibitem[He et~al.(2015)He, Zhang, Ren, and Sun]{DBLP:journals/corr/HeZRS15}
Kaiming He, Xiangyu Zhang, Shaoqing Ren, and Jian Sun.
\newblock Deep residual learning for image recognition.
\newblock \emph{CoRR}, abs/1512.03385, 2015.
\newblock URL \url{http://arxiv.org/abs/1512.03385}.

\bibitem[He et~al.(2017)He, Gkioxari, Doll{\'{a}}r, and
  Girshick]{DBLP:journals/corr/HeGDG17}
Kaiming He, Georgia Gkioxari, Piotr Doll{\'{a}}r, and Ross~B. Girshick.
\newblock Mask {R-CNN}.
\newblock \emph{CoRR}, abs/1703.06870, 2017.
\newblock URL \url{http://arxiv.org/abs/1703.06870}.

\bibitem[Jain and Sabharwal(2010)]{jain_11_bucket}
Nikhil Jain and Yogish Sabharwal.
\newblock {Optimal bucket algorithms for large MPI collectives on torus
  interconnects}.
\newblock \emph{In proceedings of the International Conference on
  SUpercomputing}, 2010.

\bibitem[Jouppi et~al.(2017)Jouppi, Young, Patil, Patterson, Agrawal, Bajwa,
  Bates, Bhatia, Boden, Borchers, Boyle, Cantin, Chao, Clark, Coriell, Daley,
  Dau, Dean, Gelb, Ghaemmaghami, Gottipati, Gulland, Hagmann, Ho, Hogberg, Hu,
  Hundt, Hurt, Ibarz, Jaffey, Jaworski, Kaplan, Khaitan, Koch, Kumar, Lacy,
  Laudon, Law, Le, Leary, Liu, Lucke, Lundin, MacKean, Maggiore, Mahony,
  Miller, Nagarajan, Narayanaswami, Ni, Nix, Norrie, Omernick, Penukonda,
  Phelps, Ross, Salek, Samadiani, Severn, Sizikov, Snelham, Souter, Steinberg,
  Swing, Tan, Thorson, Tian, Toma, Tuttle, Vasudevan, Walter, Wang, Wilcox, and
  Yoon]{Jouppi_17}
Norman~P. Jouppi, Cliff Young, Nishant Patil, David~A. Patterson, Gaurav
  Agrawal, Raminder Bajwa, Sarah Bates, Suresh Bhatia, Nan Boden, Al~Borchers,
  Rick Boyle, Pierre{-}luc Cantin, Clifford Chao, Chris Clark, Jeremy Coriell,
  Mike Daley, Matt Dau, Jeffrey Dean, Ben Gelb, Tara~Vazir Ghaemmaghami,
  Rajendra Gottipati, William Gulland, Robert Hagmann, Richard~C. Ho, Doug
  Hogberg, John Hu, Robert Hundt, Dan Hurt, Julian Ibarz, Aaron Jaffey, Alek
  Jaworski, Alexander Kaplan, Harshit Khaitan, Andy Koch, Naveen Kumar, Steve
  Lacy, James Laudon, James Law, Diemthu Le, Chris Leary, Zhuyuan Liu, Kyle
  Lucke, Alan Lundin, Gordon MacKean, Adriana Maggiore, Maire Mahony, Kieran
  Miller, Rahul Nagarajan, Ravi Narayanaswami, Ray Ni, Kathy Nix, Thomas
  Norrie, Mark Omernick, Narayana Penukonda, Andy Phelps, Jonathan Ross, Amir
  Salek, Emad Samadiani, Chris Severn, Gregory Sizikov, Matthew Snelham, Jed
  Souter, Dan Steinberg, Andy Swing, Mercedes Tan, Gregory Thorson, Bo~Tian,
  Horia Toma, Erick Tuttle, Vijay Vasudevan, Richard Walter, Walter Wang, Eric
  Wilcox, and Doe~Hyun Yoon.
\newblock In-datacenter performance analysis of a tensor processing unit.
\newblock In \emph{Proceedings of ISCA'17}, 2017.
\newblock URL \url{http://arxiv.org/abs/1704.04760}.

\bibitem[{Kumar} et~al.(2014){Kumar}, {Laxmi}, {Gaur}, {Daneshtalab},
  {Ebrahimi}, and {Zwolinski}]{on_chip_fault_tolerant_routing}
M.~{Kumar}, V.~{Laxmi}, M.~S. {Gaur}, M.~{Daneshtalab}, M.~{Ebrahimi}, and
  M.~{Zwolinski}.
\newblock Fault tolerant and highly adaptive routing for 2d nocs.
\newblock In \emph{2014 IEEE International Symposium on Defect and Fault
  Tolerance in VLSI and Nanotechnology Systems (DFT)}, pages 104--109, 2014.

\bibitem[Kumar et~al.(2019)Kumar, Bitorff, Chen, Chou, Hechtman, Lee, Kumar,
  Mattson, Wang, Wang, et~al.]{kumar2019scale}
Sameer Kumar, Victor Bitorff, Dehao Chen, Chiachen Chou, Blake Hechtman,
  HyoukJoong Lee, Naveen Kumar, Peter Mattson, Shibo Wang, Tao Wang, et~al.
\newblock Scale {MLPerf-0.6} models on {Google TPU-v3} pods.
\newblock \emph{arXiv preprint arXiv:1909.09756}, 2019.

\bibitem[Li et~al.(2020)Li, Zhao, Varma, Salpekar, Noordhuis, Li, Paszke,
  Smith, Vaughan, Damania, and Chintala]{li2020pytorch}
Shen Li, Yanli Zhao, Rohan Varma, Omkar Salpekar, Pieter Noordhuis, Teng Li,
  Adam Paszke, Jeff Smith, Brian Vaughan, Pritam Damania, and Soumith Chintala.
\newblock Pytorch distributed: Experiences on accelerating data parallel
  training, 2020.

\bibitem[Russakovsky et~al.(2015)Russakovsky, Deng, Su, Krause, Satheesh, Ma,
  Huang, Karpathy, Khosla, Bernstein, Berg, and Fei-Fei]{ILSVRC15}
Olga Russakovsky, Jia Deng, Hao Su, Jonathan Krause, Sanjeev Satheesh, Sean Ma,
  Zhiheng Huang, Andrej Karpathy, Aditya Khosla, Michael Bernstein,
  Alexander~C. Berg, and Li~Fei-Fei.
\newblock {ImageNet Large Scale Visual Recognition Challenge}.
\newblock \emph{International Journal of Computer Vision (IJCV)}, 115\penalty0
  (3):\penalty0 211--252, 2015.
\newblock \doi{10.1007/s11263-015-0816-y}.

\bibitem[van~den Oord et~al.(2013)van~den Oord, Dieleman, and
  Schrauwen]{NIPS2013_5004}
Aaron van~den Oord, Sander Dieleman, and Benjamin Schrauwen.
\newblock Deep content-based music recommendation.
\newblock In C.~J.~C. Burges, L.~Bottou, M.~Welling, Z.~Ghahramani, and K.~Q.
  Weinberger, editors, \emph{Advances in Neural Information Processing Systems
  26}, pages 2643--2651. Curran Associates, Inc., 2013.
\newblock URL
  \url{http://papers.nips.cc/paper/5004-deep-content-based-music-recommendation.pdf}.

\bibitem[Vaswani et~al.(2017)Vaswani, Shazeer, Parmar, Uszkoreit, Jones, Gomez,
  Kaiser, and Polosukhin]{DBLP:journals/corr/VaswaniSPUJGKP17}
Ashish Vaswani, Noam Shazeer, Niki Parmar, Jakob Uszkoreit, Llion Jones,
  Aidan~N. Gomez, Lukasz Kaiser, and Illia Polosukhin.
\newblock Attention is all you need.
\newblock \emph{CoRR}, abs/1706.03762, 2017.
\newblock URL \url{http://arxiv.org/abs/1706.03762}.

\bibitem[Xu et~al.(2020)Xu, Lee, Chen, Choi, Hechtman, and
  Wang]{xu2020automatic}
Yuanzhong Xu, HyoukJoong Lee, Dehao Chen, Hongjun Choi, Blake Hechtman, and
  Shibo Wang.
\newblock {Automatic Cross-Replica Sharding of Weight Update in Data-Parallel
  Training}.
\newblock \emph{arXiv preprint arXiv:2004.13336}, 2020.

\bibitem[Ying et~al.(2018)Ying, Kumar, Chen, Wang, and Cheng]{resnet_18}
Chris Ying, Sameer Kumar, Dehao Chen, Tao Wang, and Youlong Cheng.
\newblock Image classification at supercomputer scale.
\newblock \emph{CoRR}, abs/1811.06992, 2018.
\newblock URL \url{http://arxiv.org/abs/1811.06992}.

\end{thebibliography}
\bibliographystyle{plainnat}

\end{document}